\documentclass{article}
\usepackage{ijcai25}

\usepackage{times}
\usepackage{soul}
\usepackage{url}
\usepackage[hidelinks]{hyperref}
\usepackage[utf8]{inputenc}
\usepackage[small]{caption}
\usepackage{graphicx}
\usepackage{amsmath}
\usepackage{amsthm}
\usepackage{booktabs}
\usepackage[switch]{lineno}

\usepackage{amssymb}
\usepackage{xcolor}
\usepackage[ruled,vlined,linesnumbered]{algorithm2e}
\usepackage{centernot}
\usepackage{stackengine}
\usepackage{subfig}
\usepackage{multirow}
\usepackage{makecell}

\theoremstyle{definition}
\newtheorem{definition}{Definition}[section]

\title{Inference of Human-derived Specifications of Object Placement via Demonstration}

\author{
Alex Cuellar$^1$
\and
Ho Chit Siu$^2$\And
Julie A Shah$^1$
\affiliations
$^1$Massachusetts Institute of Technology\\
$^2$MIT Lincoln Laboratory\\
\emails
alexcuel@mit.edu,
julie\_a\_shah@csail.mit.edu,
hochit.siu@ll.mit.edu
}

\begin{document}

\maketitle

\begin{abstract}
As robots' manipulation capabilities improve for pick-and-place tasks (e.g., object packing, sorting, and kitting), methods focused on understanding human-acceptable object configurations remain limited expressively with regard to capturing spatial relationships important to humans. To advance robotic understanding of human rules for object arrangement, we introduce positionally-augmented RCC (PARCC), a formal logic framework based on region connection calculus (RCC) for describing the relative position of objects in space. Additionally, we introduce an inference algorithm for learning PARCC specifications via demonstrations. Finally, we present the results from a human study, which demonstrate our framework's ability to capture a human's intended specification and the benefits of learning from demonstration approaches over human-provided specifications. 

\end{abstract}
\section{Introduction}

As robots become a mainstay of industrial and manufacturing processes, pick-and-place tasks (e.g., packing, sorting, and kitting objects) have become central to many of their applications \cite{state-of-robots}. While significant prior research has explored control and manipulation for pick-and-place tasks, less has examined how robots can understand human preferences about object arrangement. Erbayrak et al, for example, designed an algorithm to pack objects into multiple bins while optimizing to keep as many objects of the same predefined ``family'' together as possible \cite{erbayrak2021multi}. Sun et al studied algorithms designed to instruct warehouse workers where to place items in a box, and observed when workers knowingly deviated from the algorithm's plan \cite{sun2022predicting}. The researchers then proposed modified algorithms to minimize human deviation from plans. While such approaches show promise in specific domains, they have limited expressiveness for required spatial relationships during object placement tasks. Therefore, in this work we present Positionally-Augmented Region Connection Calculus (PARCC), a spatial specification language able to capture humans' requirements during object placement tasks. Additionally, we introduce an inference algorithm to infer PARCC specifications from demonstrations. 

\begin{figure}
    \centering
    \includegraphics[width=1\columnwidth]{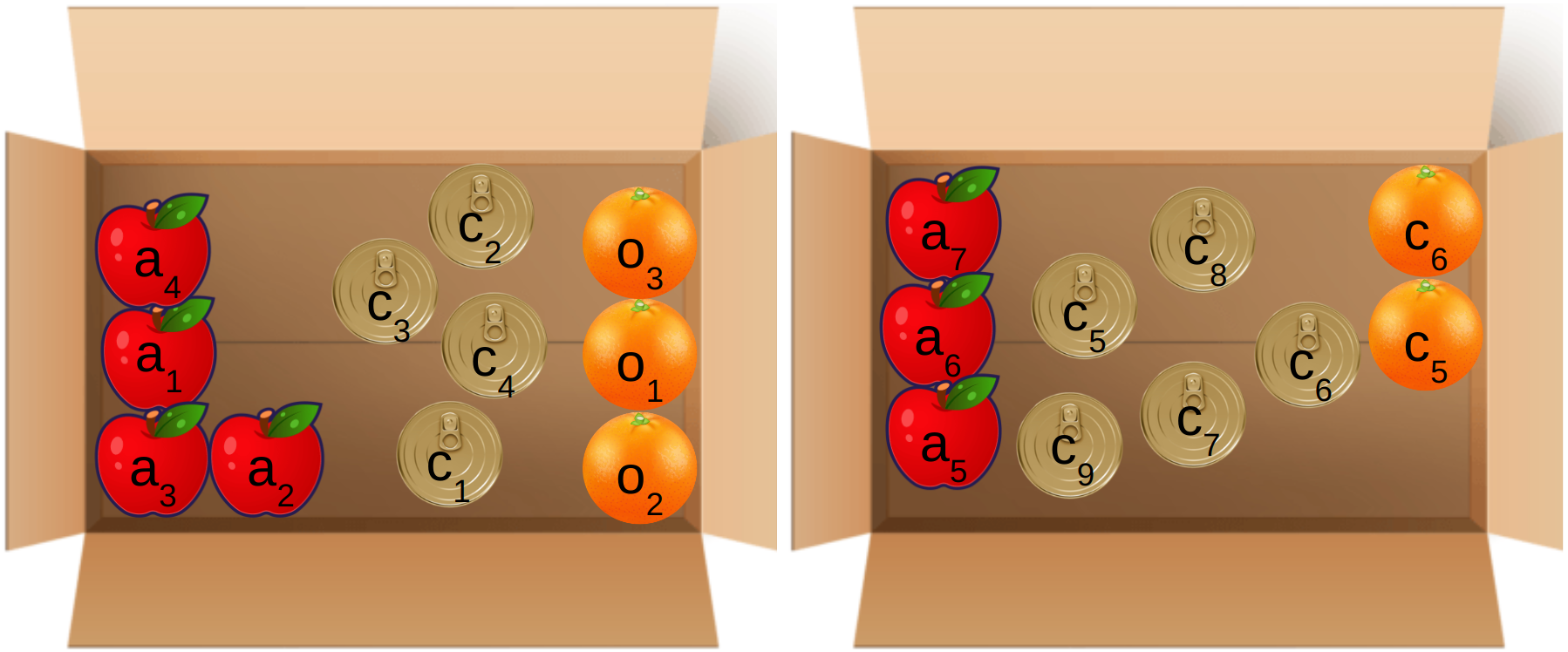}
    \caption{Two example configurations of apples, oranges, and cans in a box.}
    \label{fig:PARCC_Illustrative_Examples}
\end{figure}

Broadly speaking, capturing humans' understanding of objects' spatial relationships in a scene is not a new topic to research. Paul et al introduced a framework to ``ground" human instructions in a world representation (e.g., understanding the instruction ``pick up the middle block in the row of 5 blocks") \cite{paul2018efficient}, while others have developed methods to generate scenes or object arrangements from descriptions, either via predefined propositions (e.g., ``A is left of B") \cite{wiebrock2000inference} or through natural language \cite{vasardani2013descriptions,liu2022structformer}. 

While these systems could theoretically be used by workers to communicate specifications, there are a few underlying limitations. First, these methods only describe one scene at a time: for example, in Figure \ref{fig:PARCC_Illustrative_Examples}, prior methods may be able to express that the can $c_2$ is the furthest-north object in its scene, or generate an approximation of a scene via description of each object's placement. However, when considering a specification describing both scenes, these methods cannot communicate concepts such as ``all cans are east of oranges" or ``all oranges must touch another orange to the north or south." Capturing these relational 
rules can ensure, for example, that more fragile objects like apples or oranges are properly supported via contact or that there is consistent spatial placing of objects that may be expected by human workers. Additionally, methods capturing a human’s understanding of a scene rely upon the human directly providing a description of that scene \cite{wiebrock2000inference,vasardani2013descriptions,liu2022structformer}. However, humans often under-specify (or misspecify) tasks, relative to what robots require, leading to undesired behavior when following a human's directly-provided specification \cite{gross2016multi}.
 
To address such limitations, PARCC is designed to both capture descriptions over a ``class" of related objects (e.g., apples, oranges, or cans) and use Boolean logic to encode specifications (e.g., ``all oranges must touch another orange to the north or south"). Additionally, we present an inference framework to infer specifications from demonstrations instead of relying upon human-provided specifications. To demonstrate the effectiveness of our framework, we performed a human study to test how well the inference method captured specifications via demonstrations compared with direct human-provided specifications.  In the field, this method can fit into a larger pipeline for automation of object placement tasks including packing and sorting that maintains patters of human behavior.  Continuing the box packing example from Figure \ref{fig:PARCC_Illustrative_Examples}, a human can demonstrate multiple examples of packing apples, cans, and oranges into boxes and a PARCC specification of object relations can be inferred.  In the future, a robot performing the packing task can plan and execute object placements satisfying the demonstrated specification. 

\section{Related Works}
\label{RelatedWorks}


Within the field of task specification via formal logics, descriptions of spatial specifications often use Signal Temporal Logic (STL), since its operation over continuous signals makes it ideal for expressing spatial preferences \cite{maler2004monitoring}. Application of STL to spatial problems either uses standard STL operators to express positions and regions as signals \cite{linard2020active} or modifies notation to include spatial-specific operators \cite{nenzi2015qualitative,ma2020sastl}. For example, Nenzi et al added two spatial modalities (``somewhere" and ``surrounds") to STL that operate over an undirected graph representing space (e.g., $\phi_1 \mathcal{S}_{[d_1,d_2]} \phi_2$ expresses that a region where $\phi_1$ is true is surrounded by the region where $\phi_2$ is true) \cite{nenzi2015qualitative}. 

Other spatial specification languages use quad-tree representations \cite{haghighi2015spatel}.  Here, space is recursively partitioned into quadrants over which a specification reasons.  For example, a quad-tree may represent a city and the specification describes power grid requirements across the city.

While STL and quad-tree representations are highly expressive, they do not lend themselves to cleanly encoding human-intuitive spatial relationships between objects, which tend to use qualitative descriptors and small, countable values. Conversely, region connection calculus (RCC) is a spatial-relational language introduced by Randel et al to formalize human-intuitive concepts of spatial relationships between regions \cite{randell1992spatial}. The fundamental relation of RCC is $C(x, y)$ — read as `$x$ connects with $y$', meaning that the topological closure of regions $x$ and $y$ share at least one point. The two axioms for C are as follows: 
\begin{align}
    &\forall x [C(x,x)] \\ 
    &\forall x \forall y [C(x,y) \rightarrow C(y,x)]
\end{align}
The first axiom states that any region $x$ must connect to itself; the second states that if $y$ connects to $x$, $x$ must connect to $y$. This fundamental ``connect'' relation can be used to describe many spatial relations, and several fragments have been proposed for various purposes; RCC8, for example, is a set of eight exhaustive and pairwise disjoint relationships within the RCC framework. However, in this paper we focus on the original 10 relations described in Technical Appendix A. Technical Appendix B further discusses differences in expressing qualitative object relations using PARCC, STL, and quad-trees; as an extension of RCC, PARCC expresses objects relations more easily than other languages.

While RCC itself is not based in logical specification, it has been adopted into specification languages. Ven et al introduced qualitative privacy description language (QPDL), using RCC within linear temporal logic (LTL) to describe technological privacy \cite{van2016qualitative}. Similarly, spatio-temporal synthesis logic (STSL) combines $SU_4$ (a spatial language similar to RCC) with STL to characterize spatio-temporal dynamic behaviors in applications such as adaptive cruise control \cite{li2020stsl}. However, neither QPDL nor STSL can express specification over ``classes" of regions (i.e., ``oranges are east of cans," as shown in Figure \ref{fig:PARCC_Illustrative_Examples}), nor has an algorithm been proposed to infer such specifications from demonstration.  


With respect to specification inference via demonstrations, we take inspiration from Vazquez et al, who proposed an inference framework over LTL via a maximum entropy approach \cite{vazquez2018learning}. This framework derives a likelihood model over specifications based on how frequently demonstrations satisfy a specification versus the probability that the specification would be satisfied by random actions. 


\section{PARCC Formulation}
\label{problem_formulation}

Applying RCC directly to describe object relations provides some insight into human-intuitive relationships between objects. For example, in Figure \ref{fig:PARCC_Illustrative_Examples}, RCC can describe that $a_1$ is in contact with $a_4$ using the notation $EC(a_1, a_4)$; however, RCC fails to capture two aspects of the examples in Figure \ref{fig:PARCC_Illustrative_Examples}. First, RCC cannot communicate directionality: while a human may naturally notice that $a_1$ is west of $c_1$ or that $o_3$ contacts $o_1$ on the north side, RCC cannot capture these distinctions. Second, while RCC describes the relationship between individual objects (e.g., $EC(o_1, o_3)$), it fails to describe a pattern over a ``class" of similar objects (e.g. apples are west cans). 

In order to include these two capabilities into PARCC, we first define a subset of RCC relations useful for describing object relationships (as opposed to abstract regions), and augment this subset to capture directional information. We then use these object relations to describe patterns between all objects of a particular class using Boolean logic. 

We constrain PARCC to exist over axis-aligned rectangular objects on a flat plane; this choice is motivated by the ubiquitous use of rectangular bounding boxes to represent the size and location of objects in a scene \cite{ali2024yolo,zendehdel2023real,he2021know,jia2021self}. In this paper, we define an object as a tuple, $o = (o_l, o_w, o_x, o_y, o_c)$, where $o_l$ is the object's length, $o_w$ is its height, $o_x$ and $o_y$ designate its position in space, and $o_c$ is the object's ``class" — a label provided to objects subject to the same specifications (e.g., apples, oranges, and cans in Figure \ref{fig:PARCC_Illustrative_Examples}). In application, this could designate fragile vs. non-fragile objects, shipping destinations for packages, etc. 

\subsection{PARCC Relations}
\label{Relations}

PARCC reasons over object relations via a subset of the canonical RCC relations (Technical Appendix A) --- specifically with ``discrete from” and ``externally connected to” (written as $DR(x, y)$ and $EC(x, y)$, respectively). As described in Technical Appendix A, $DR(x, y)$ implies the interiors of $x$ and $y$ do not overlap, and $EC(x, y)$ implies the exterior boundaries $x$ and $y$ touch. 
We exclude the remaining RCC relations since they describe some overlap between regions (disallowed as our regions represent physical objects) or can be described with $DR$ and $EC$ themselves.

\begin{definition}[PARCC object relation]
    PARCC object relations include the basic $DR$ and $EC$ relations, with a subscript indicating the relative cardinal position of one object to another (assuming north is aligned with the positive $y$ axis). For example, in Figure \ref{fig:PARCC_Illustrative_Examples}, we can say that $DR_N(c_2, c_1)$; formally, this requires the following:
\begin{multline}
    \label{PositionAugmentedRelation}
    DR_N(c_2, c_1) \rightarrow DR(c_2, c_1) \land y_{c_2} \geq y_{c_1} \quad \\ \forall (x_{c_2}, y_{c_2}) \in c_2, (x_{c_1}, y_{c_1}) \in c_1
\end{multline}


\end{definition}

meaning $c_2$ is discrete from $c_1$, and the $y$ value of every point in $c_2$ is greater than the $y$ value of every point in $c_1$. \\

In addition to the position-augmented object relations, our language must also reason over relations between classes. For this purpose, we define class relations as follows: 

\begin{definition}[PARCC class relation]
    \label{Definition:PARCCClassRElation}
    PARCC class relations use the same notation as position-augmented object relations, but operate over two classes. Given that $\mathcal{A}$ is the set of objects in class $A$ and $\mathcal{B}$ is the set of objects in class $B$, our class $DR$ relations require all objects in $\mathcal{A}$ to have the provided relation with all objects in $\mathcal{B}$. For example, a $DR$ North class relation would be as follows:
    \begin{align}
    \label{DRClassRel}
            DR_N(A, B) \leftrightarrow DR_N(a, b) \quad \forall a \in \mathcal{A} \quad \forall b \in \mathcal{B}
    \end{align}
    This means that, for all objects $a$ of class $A$ and all objects $b$ of class $B$, $DR_N(a, b)$ must hold. Conversely, $EC$ class relations would require that all objects in $\mathcal{A}$ have the given position augmented relation with at least one object in $\mathcal{B}$. For example, a $EC$ North class relation would be as follows:

    \begin{align}
    \label{ECClassRel}
            EC_N(A, B) \leftrightarrow EC_N(a, b) \quad \forall a \in \mathcal{A} \quad \exists b \in \mathcal{B}
    \end{align}

This means that, for all objects $a$ of class $A$, there exists an object $b$ of class $B$ such that $EC_N(a, b)$. 
    \end{definition}
    
\subsubsection{PARCC Formulas}
\label{PARCCformulas}

Prior work has used RCC relations as propositions in logic languages \cite{van2016qualitative}; we extend this to Boolean logic over PARCC class relations. 

\begin{definition}[PARCC Formula]
    A PARCC formula is a propositional logic formula over PARCC class relations. Conjunction and disjunction over PARCC class relations can be applied directly using the definitions of class relations in Eqs \ref{DRClassRel} and \ref{ECClassRel}. For example: 
\begin{multline}
    DR_N(A, B) \lor DR_S(C, D) \leftrightarrow \\ (DR_N(a, b) \forall a \in \mathcal{A} \, \forall b \in \mathcal{B}) \, \lor \, (DR_S(c, d) \forall c \in \mathcal{C} \, \forall d \in \mathcal{D})
\end{multline}
    Negation of $DR$ and $EC$ PARCC class relations are defined as follows: 
\begin{align}
        \lnot DR_i(A, B) \leftrightarrow \nexists a \in \mathcal{A} \quad s.t. \;  DR_i(a, b) \; \forall b \in \mathcal{B} \\
        \lnot EC_i(A, B) \leftrightarrow \nexists a \in \mathcal{A} \quad s.t.\; EC_i(a, b) \; \exists b \in \mathcal{B}   
\end{align}
In order for $\lnot DR_i(A, B)$ to hold, there cannot exist an object $a$ of class $A$ for which $DR_i(a, b)$ holds for all objects of class $B$. Conversely, for $\lnot EC_i(A, B)$ to hold, there cannot exist an object $a$ of class $A$ for which $EC_i(a, b)$ holds for any objects of class $B$. 

\end{definition}

\subsubsection{Demonstration}
\label{SpatialContext}
In this paper, we describe our inference algorithm over ``demonstrations” of objects in a given space.

\begin{definition}[Demonstration]
    We define a demonstration as the tuple $D = (\mathcal{O}_D, \mathcal{S}_D, \mathcal{L}_D)$, where $\mathcal{O}_D$ is the set of objects in a particular demonstration, $\mathcal{S}_D$ is the space of $x$, $y$ points available for object placement (i.e., $(o_x, o_y) \in \mathcal{S}_D \forall o \in \mathcal{O}_D$), and $\mathcal{L}_D$ is the set of classes to which objects in $\mathcal{O}_D$ belong. We notate $\mathcal{O}_D^L$ as the subset of objects belonging to class $L \in \mathcal{L}$. 
\end{definition}

Throughout this paper, we will use PARCC formulas to describe a demonstration, $D$. Specifically, we say $D$ can satisfy a formula, $\phi$, if all objects in $\mathcal{O}_D$ satisfy $\phi$. For example, let $D$ be the demonstration of the example on the left in Figure $\ref{fig:PARCC_Illustrative_Examples}$, $A$ be the class of apples, and $C$ be the class of cans. Statement $D \rightarrow DR_E(A, C)$ then evaluates to true, since every apple is discrete from and east of every can. However, the statement $D \rightarrow EC_E(A, A)$ evaluates to false, since not all apples are externally connected to another apple to the east. 


\section{Specification Inference}
\label{SpecificationInference}

For our inference procedure, we assume access to $k$ demonstrations from a human, notated $\mathcal{D} = \{D_1 ... D_k\}$. We assume each $D_i$ has the same space $\mathcal{S}_D$ and classes $\mathcal{L}_D$; note, however, that demonstrations have different object sets, $\mathcal{O}_D$. We also assume each demonstration $D \in \mathcal{D}$ conforms to a specification $\Phi_h$. Our goal is to infer a conjunctive normal-form (CNF) PARCC formula, $\Phi$, that describes universal patterns in $D_1 .... D_k$ as intended by the demonstrator. 

The inference process has two steps. First, we use a search-based method to determine a set of disjunctive PARCC formulas, $\Bar{\mathcal{C}}$, such that for each $\phi \in \Bar{\mathcal{C}}$, $D \rightarrow \phi$ for all $D \in \mathcal{D}$; these will form the disjunctive clauses of the CNF formula, $\Phi$. Second, using a frequentist approach, we determine the probability that each formula $\phi \in \Bar{\mathcal{C}}$ was intended by the demonstrator.  We evaluate this probability using $\mathcal{R}$, a set of ``non-specification" demonstrations generated without the human's specification, and calculate the probability that a demonstration would satisfy $\phi$ without intent. 

\begin{algorithm}[t]
 \caption{Candidate Disjunctive Formulas}\label{CandidateAlg}
  \SetKwFunction{FInference}{FindDisjunctions}
  \SetKwProg{Fn}{Function}{:}{}
  \Fn{\FInference{$D_1...D_d$, $\mathcal{L}$, $N$}}{
    $\Bar{\mathcal{C}} = \emptyset$ \;
    \For{n = 1 ... N}{
         \For{$\phi \in Template(\mathcal{L}, n)$}{
                \If{$\exists \Bar{\phi} \in \Bar{\mathcal{C}} \; s.t. \; \Bar{\phi} \rightarrow \phi$}{
                    continue \;
                }
                \If{$D \rightarrow \phi \; \forall \; D \in \{D_1.... D_d\}$}{
                    $\Bar{\mathcal{C}}.add(\phi)$ \;
                }
            }
        }
        \KwRet{$\Bar{\mathcal{C}}$}
  }
\end{algorithm}

\subsection{Finding Satisfying Disjunctive Formulas}
\label{FindingCandidateConstraints}

Our inference procedure begins by finding a set of disjunctive PARCC formulas, $\Bar{\mathcal{C}}$, which satisfy all demonstrations, $\mathcal{D}$ (Algorithm \ref{CandidateAlg}). We use an exhaustive search-based method to determine $\Bar{\mathcal{C}}$. While one can brute-force a search over all possible disjunctive formulas, this would require checking $\mathcal{O}(|\mathcal{L}|^{2N}4^N)$ formulas.  Therefore, we take inspiration from Shah et al., and allow the use of production rule ``templates" limiting the search space based on domain knowledge (see Section \ref{subsec:TemplateChoice} for an example) \cite{shah2018bayesian}. 

This procedure takes in demonstrations ($\mathcal{D}$) the set of object classes ($\mathcal{L}$) and the maximum length of the disjunctive formulas (N). First, the algorithm initializes the set of disjunctive formulas, $\Bar{\mathcal{C}}$, as the empty set (line 2); we then loop over all possible lengths of the disjunctive phrase from 1 to $N$ (line 3). For each length, we loop over every disjunctive formula of length $n$ allowed by the production rule template given the classes, $\mathcal{L}$ (line 4). For each formula $\phi$, we check whether a disjunction is already trivially implied by another formula in $\Bar{\mathcal{C}}$ — and, if so, do not further consider it for $\Bar{\mathcal{C}}$ (lines 5-6). Next, we check if $\phi$ is satisfied by all demonstrations in $\mathcal{D}$, and if so add it to $\Bar{\mathcal{C}}$ (lines 7-8). Finally, we return the satisfying formulas, $\Bar{\mathcal{C}}$ (line 9). 

\subsection{Determining Intended Disjunctive Formulas}
\label{DeterminingCandidateConstraintRelevance}
Once our algorithm finds $\Bar{\mathcal{C}}$, we determine the formulas $\phi \in \Bar{\mathcal{C}}$ that likely describe the demonstrator’s intent. To this end, we calculate the probability that any demonstration $D$ will unintentionally satisfy $\phi$, which we express as $P(D \rightarrow \phi | \mathcal{R})$, where $\mathcal{R}$ is a set of ``non-specification" demonstrations generated without the human's intended specification, $\Phi_h$. 

Notice that $D \rightarrow \phi$ if and only if the relations every object $o$ has with other objects in the demonstration satisfy $\phi$ (which we will notate as $o \rightarrow \phi$). Therefore, assuming the probability that each object satisfies $\phi$ is independent, the probability is as follows, where we use $C$ to notate the class of objects relevant to $\phi$: 
\begin{align}
\label{ProbabilyOfDemo}
    P(D \rightarrow \phi | \mathcal{R}) = \prod_{o \in \mathcal{O}_D^{C}} P(o \rightarrow \phi | \mathcal{R})
\end{align}
While not necessarily representative of reality, our independence assumption makes this probabilistic modeling tractable, and provides good results in our human study (see Section \ref{Evaluation}). In order to approximate $p(o \rightarrow \phi | \mathcal{R})$, we calculate the fraction of objects from $\mathcal{R}$ that satisfy formula $\phi$: 
\begin{multline}
\label{ProbabilityOfObject}
    P(o \rightarrow \phi | \mathcal{R}) = \\
    \max \left(\epsilon, \frac{\sum_{R \in \mathcal{R}} \sum_{o' \in \mathcal{O}_{R}^C}\mathbf{1}(o' \rightarrow \phi)}{\sum_{R \in \mathcal{R}} \sum_{o' \in \mathcal{O}_{R}^C} 1} \right)
\end{multline}
\noindent where $\epsilon$ is a small number that prevents the probability from being 0 (we use .01), and $\mathbf{1}(o' \rightarrow \phi)$ is an indicator variable set to 1 if $o' \rightarrow \phi$, and 0 otherwise. We substitute this probability back into Equation \ref{ProbabilyOfDemo}, which results in the following:
\begin{multline}
    \label{InferenceEq}
    P(D \rightarrow \phi | \mathcal{R}) = \\\prod_{o \in \mathcal{O}_D^{C}}  \max \left(\epsilon, \frac{\sum_{R \in \mathcal{R}} \sum_{o' \in \mathcal{O}_{R}^C}\mathbf{1}(o' \rightarrow \phi)}{\sum_{R \in \mathcal{R}} \sum_{o' \in \mathcal{O}_{R}^C} 1} \right)
\end{multline}
Algorithm \ref{InferAlg} describes the construction of $\Phi$. The algorithm takes a set of human demonstrations ($\mathcal{D}$), the number of non-specification demonstrations to generate ($k_r$), the set of disjunctive formulas found in Algorithm \ref{CandidateAlg} ($\Bar{\mathcal{C}}$), and a cutoff probability parameter ($p_c$). First, the algorithm constructs an empty set, $\mathcal{C}$ (line 2).  Then the algorithm generates $k_r$ non-specification demonstrations via the \verb|SampleRandDemo| function (line 3). \verb|SampleRandDemo| copies a demonstration from $\mathcal{D}$ (line 10) and reassigns each object a point in the demonstration space, $\mathcal{S}_D$ (line 12). Once every object's position is reassigned, the demonstration is returned (line 13). (While the pseudo-random object placement in \verb|SampleRandDemo| may be distinct from human behavior without a defined $\Phi_h$, experiments presented in Section \ref{Evaluation} show the process described here provides a reasonable analog to human-provided non-specification demonstrations.)

\begin{algorithm}[t]
 \caption{Inferring Intended Formulas} \label{InferAlg}
  \SetKwFunction{FInference}{Inference}
  \SetKwFunction{FSample}{SampleRandDemo}
  \SetKwProg{Fn}{Function}{:}{}
  \Fn{\FInference{$\mathcal{D}$, $\Bar{\mathcal{C}}$, $p_c$, $k_r$}}{
    $\mathcal{C} = \emptyset$ \\
    $\mathcal{R} = \{ $ \FSample{$\mathcal{D}$} $ | i \in k_r \}$ \\
    \For{$\phi \in \Bar{\mathcal{C}}$}{
        $p_\phi = \prod_{D \in \mathcal{D}} P(D \rightarrow \phi | \mathcal{R})$ \\
        \If{$p_\phi < p_c$}{
            $\mathcal{C}.add(\phi)$
        }
    }
    \KwRet $\bigwedge_{\phi \in \mathcal{C}} \phi$
  }
  \SetKwProg{Fn}{Function}{:}{}
  \Fn{\FSample{$\mathcal{D}$}}{
    $R = ChooseRandom(\mathcal{D})$.copy() \\
    \For{$o \in \mathcal{O}_{R}$}{
        $(o_x, o_y) = ChooseRandom(\mathcal{S}_{D})$ \\
    }
    \KwRet $R$
  }    
\end{algorithm}

The algorithm then loops over every disjunctive formula $\phi \in \Bar{\mathcal{C}}$, calculating the probability that all human demonstrations unintentionally satisfied $\phi$ using Equation \ref{InferenceEq} (lines 4-5). Next, the algorithm checks whether this probability is under the cutoff probability $p_c$ (i.e., whether we are confident that $\phi$ was not randomly satisfied, we use $p_c=.05$), and adds it to $\mathcal{C}$ (lines 6-7) if so. Finally, the algorithm returns the full specification as the conjunction of all formulas in $\mathcal{C}$ (line 8).

\section{Experiment}
\label{Evaluation}

We implemented the PARCC inference procedure and evaluated it with human subjects. This study tested the effectiveness of the PARCC specification language and inference framework to capture a human’s intent in specifying a spatial configuration. Additionally, our experiment provides support for the hypothesis that demonstration-based specification systems can mitigate issues related to under- or misspecification arising from other modalities, such as natural language. We do not compare against other specification languages common in spatial domains (e.g. STL) as they are ill-equipped to represent object class relations (see Appendix B). \footnote{For code and datasets: https://github.com/AlexCuellar/PARCC}

\subsection{Experimental Setup}
\label{subsec:experimental_setup}


\begin{figure}
    \centering
    \includegraphics[width=\columnwidth]{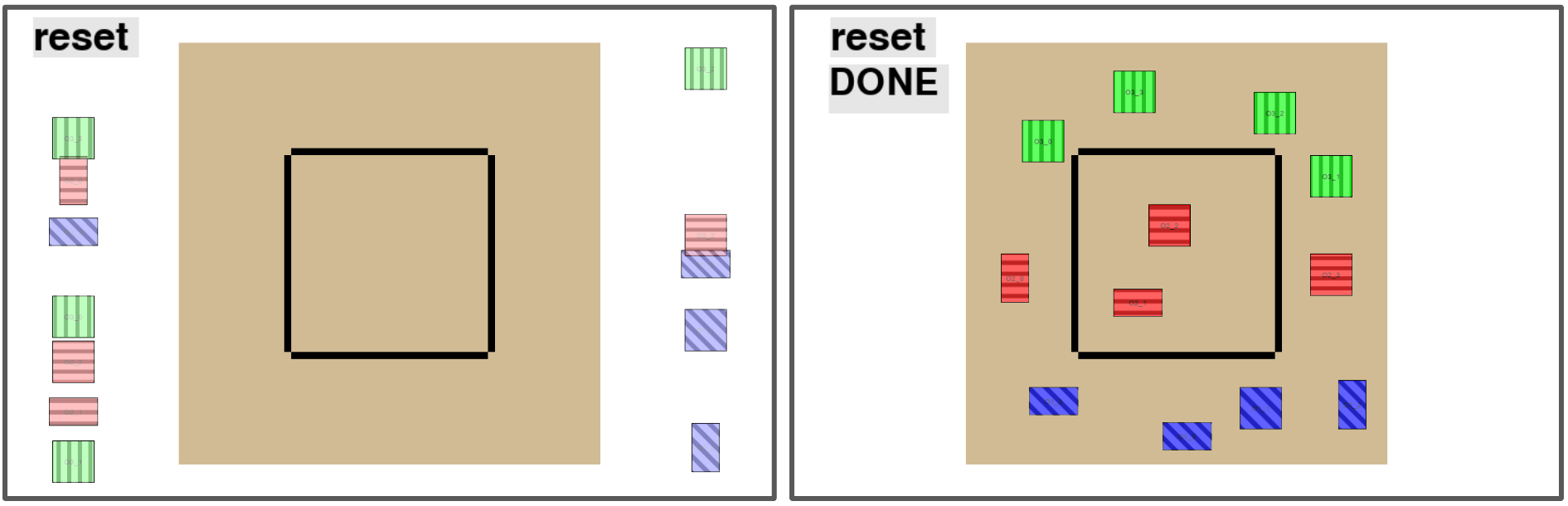}
    \caption{The demonstration interface we used in our experiment.  The initial state (left) and completed state (right) is shown.}
    \label{fig:Initial_and_Final_Interface}
\end{figure}

For this study, we designed a box-packing environment as a representative task (Figure \ref{fig:Initial_and_Final_Interface}). The environment initializes with a brown square representing a table acting as the demonstration space, and four objects representing walls of an open box. Off the table were two to four objects each of classes $R$ (red objects), $G$ (green objects), and $B$ (blue objects), which the subject could move. For visual distinction in figures, we shade red objects with horizontal lines, green objects with vertical lines, and blue objects with diagonal lines. Once the subject placed all objects on the table, a ``done" button appeared, allowing the subject to complete the demonstration.

\begin{figure}
    \centering
    \includegraphics[width=\columnwidth]{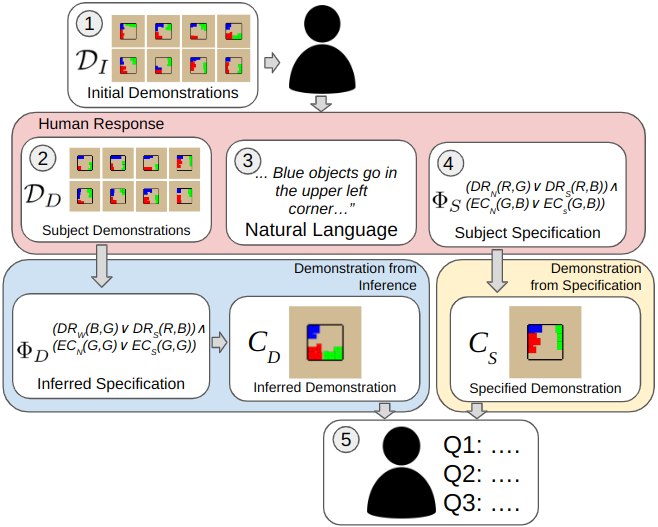}
    \caption{A pipeline showing our human study procedure. Steps directly involving human participation are numbered 1-5. Section \ref{subsec:experimental_setup} depicts the final questions given to the human.}
    \label{fig:study_pipeline}
\end{figure}

\begin{figure}
    \centering
    \includegraphics[width=\columnwidth]{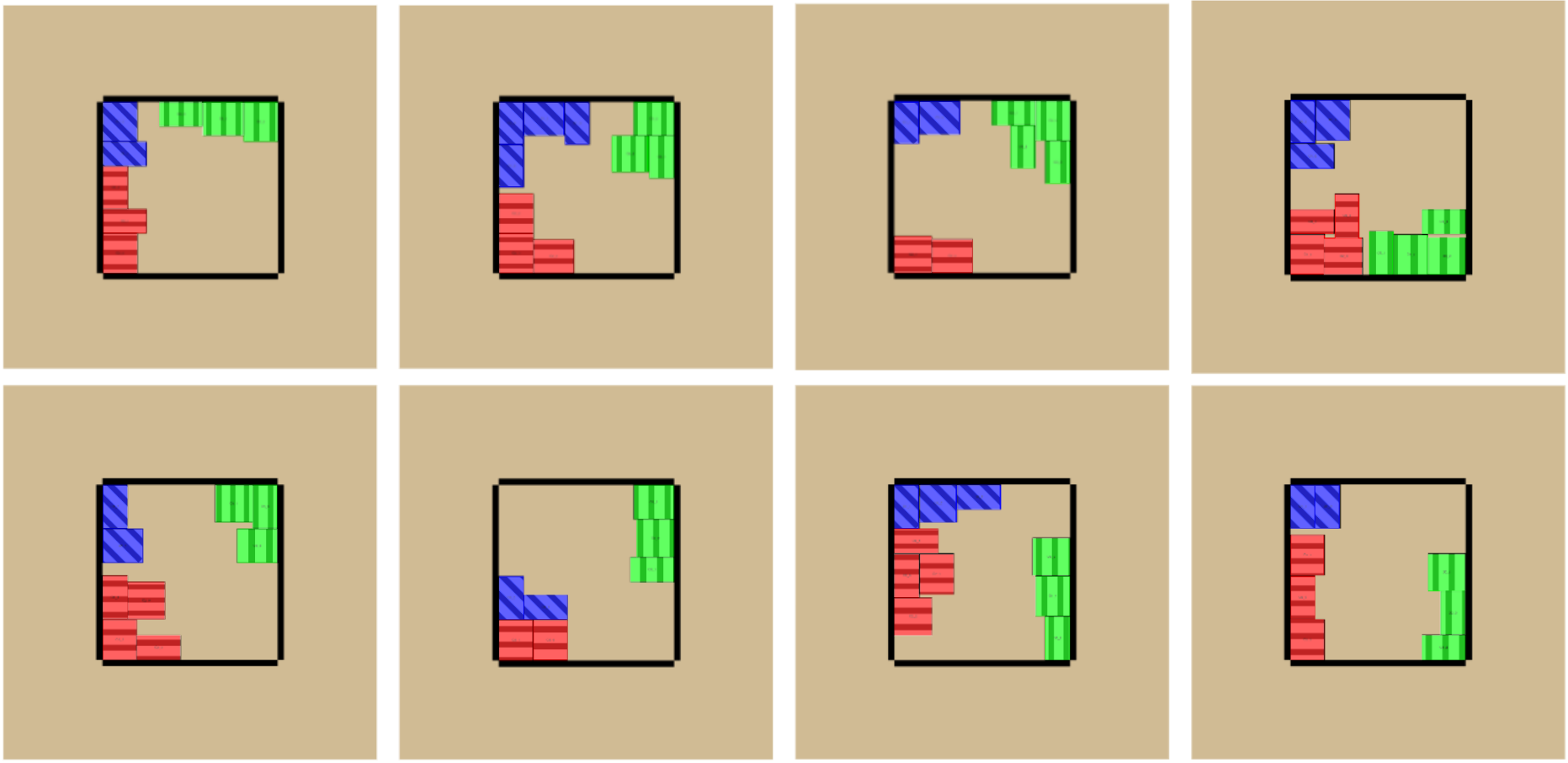}
    \caption{Pre-generated demonstrations of the box packing environment initially shown to study subjects ($\mathcal{D}_I$).}
    \label{fig:packing_examples}
\end{figure}

The experiment procedure began with a participant training phase, during which we asked the participant to provide five object placement demonstrations (training demonstrations $\mathcal{D}_T$) without having received any prompting with regard to how to arrange the objects on the table. The remainder of the process is shown in Figure \ref{fig:study_pipeline}, along with the notation for demonstrations and specifications generated throughout the experiment. After the training, we showed each participant a set of eight pre-generated demonstrations $\mathcal{D}_I$ (Figure \ref{fig:packing_examples}). The set of pre-generated demonstrations, $\mathcal{D}_I$, were identical for all participants and were created to satisfy a conjunction of of 12 PARCC formulas described in Technical Appendix C. Upon showing participants $\mathcal{D}_I$, we requested three responses: a set of in-kind demonstrations, and two specifications. 


First, the subject provided eight demonstrations, $\mathcal{D}_D$, attempting to follow all spatial patterns they observed in $\mathcal{D}_I$. (Note we had not yet introduced the PARCC language to participants, and they were free to consider ``patterns” in whatever representation was most natural.) The participant then provided a natural-language explanation of spatial patterns in $\mathcal{D}_D$ (again, the form of these ``patterns” were determined entirely by the participant). Finally, we introduced participants to the PARCC language, and had them provide a PARCC specification $\Phi_S$ that best described their demonstrations, $\mathcal{D}_D$. Our algorithm then generated two demonstrations: $C_D$, which optimized satisfaction of the specification $\Phi_D$ inferred from $\mathcal{D}_D$; and $C_S$, which optimized the specification $\Phi_S$ provided directly by the user. These computer-generated demonstrations allowed us to probe subjects' response to system behavior when following specifications inferred from demonstrations ($\Phi_D$) versus those explicitly specified ($\Phi_S$). To generate $C_D$ and $C_S$, we used a combination of mixed-integer programming and Monte Carlo tree search, both of which have been employed in box-packing domains \cite{erbayrak2021multi,edelkamp2014monte}. Finally, we showed the participant both computer-generated demonstrations simultaneously, $C_D$ on the left and $C_S$ on the right, and asked the following Likert-style questions: 

\noindent
Q1: \textit{The left image matches patterns in my demonstrations.} \\
Q2: \textit{The right image matches patterns in my demonstrations.} \\
Q3: \textit{Which computer generated demonstration do you think better matches patterns in your demonstrations?} 

Questions were on a 1-5 scale. For Q1 and Q2, 1 indicated ``strongly disagree" and 5 indicated ``strongly agree;” for Q3, 1 indicated ``strongly left" and 5 indicated ``strongly right." 

We performed this study in two groups for whom the inference procedure used different datasets for non-specification demonstrations (notated as $\mathcal{R}$ in Section \ref{SpecificationInference}). \textit{Group A}'s specifications were inferred using the \verb|SampleRandDemo| process (Algorithm \ref{InferAlg}); \textit{Group B}'s inference used the demonstrations $\mathcal{D}_T$ provided by subjects in Group A during environment training as non-specification demonstrations $\mathcal{R}$. For each group, inference used 100 non-specification demonstrations. Via this distinction, we can identify whether using random object placements as non-specification demonstrations (see Algorithm \ref{InferAlg}) provides a good enough analog for human demonstrations without prompting for a specification. 

We recruited 35 subjects, 20 in Group A and 15 in Group B. All were aged between 18 and 34 years and had at least a high-school education. The protocol was approved by the MIT Committee on the Use of Humans as Experimental Subjects (protocol E-3748) and the United States Department of Defense Human Research Protection Office (protocol MITL20220002). All subjects provided informed consent before the experiment, and received \$15 for their participation.

\subsection{Choice of Search Template}
\label{subsec:TemplateChoice}

For inference, we allow use of a ``template" limiting the possible disjunctive formulas based on domain knowledge (Section \ref{FindingCandidateConstraints}). For this experiment, we intuited that class relations within disjunctive formulas would (1) share the same ``related" class (i.e., the first class in a PARCC class relation) and (2) always contain either $EC$ or $DC$ class relations, but not both within the same disjunction. This intuition comes from considering each disjunctive phrase as its own ``constraint," all of which must be satisfied by the overall conjunction. In this mindset, we found it likely that each ``constraint" would reason over one class and only consider one RCC relation. 

Similar to Dwyer's identification of LTL templates applicable to many real-world tasks, this template is intended to limit the search space to the most natural ``constraints" for human operators across domains \cite{dwyer1999patterns}. While a full investigation of template choice can constitute its own work, we compare our choice of template to two others in Technical Appendix D.  One template is less restrictive than the one we use (i.e. allowing a larger set of possible disjunctive formulas) and one is more restrictive (i.e. allowing a smaller set of disjunctive formulas).  The specifications inferred from human demonstrations ($\Phi_D$) are identical between our choice of template and the less restrictive template, showing that our template imposes minimal inductive bias on the specifications inferred in this experiment, and provides evidence that our template represents rules relevant to human demonstrators more generally.  Additionally, we compare our template to a more restrictive template to show how operators can intentionally eliminate rules irrelevant to a particular domain.

\subsection{Hypotheses}

To evaluate the inference framework's ability to capture humans' specification, we proposed four hypotheses.  

First, we expected that the inferred specification, $\Phi_D$, would capture a human's intended specification, and therefore hypothesized that participants would respond to (Q1) by asserting that $C_D$ matched their demonstrations: \\
\textit{\textbf{H1:} Participants agree $C_D$ matches patterns in $\mathcal{D}_D$. (Q1)} \\
Next, we expected that participants' provided specification, $\Phi_S$, would not capture their intended specification. Therefore, we hypothesized that participants would respond to (Q2) asserting that $C_S$ did not match their demonstrations: \\
\textit{\textbf{H2:} Participants disagree $C_S$ matches patterns in $\mathcal{D}_D$. (Q2)} \\
In our inference procedure (Algorithm \ref{InferAlg}), we automatically generate non-specification data $\mathcal{R}$; we expected this process to provide a reasonable analog to human demonstrations without prompting a specification. Therefore, we hypothesized that Groups A and B would respond similarly to (Q1): \\
\textit{\textbf{H3:} Response to Q1 does not significantly vary between groups A and B.} \\
Finally, we expected direct specifications to differ from inferred specifications due to under- or misspecification. \\ 
\textit{\textbf{H4:} Inferred specifications $\Phi_D$ are distinct from human-provided specifications $\Phi_S$. } 

\subsection{Results}

\begin{table*}
\centering
\begin{tabular}{ |p{4.3cm}|p{.5cm}|p{.5cm}|p{.5cm}|p{.5cm}|p{.5cm}|p{.5cm}|p{.5cm}|p{.5cm}|p{.5cm}|p{.5cm}|p{.5cm}|p{.5cm}| }
 \hline
 \multicolumn{13}{|c|}{Inclusion of Initial Specification in Response Mode} \\
\Xhline{3\arrayrulewidth}
 & $\phi_1$ & $\phi_2$ & $\phi_3$ & $\phi_4$ & $\phi_5$ & $\phi_6$ & $\phi_7$ & $\phi_8$ & $\phi_9$ & $\phi_{10}$ & $\phi_{11}$ & $\phi_{12}$ \\
\Xhline{3\arrayrulewidth}
 $\Phi_D$ (Human Demonstrations) & 1 & 1 & 1 & 1 & 1 & 1 & .88 & .91 & .91 & .77 & .74 & .88 \\
 \hline
 $\Phi_S$ (Human Specification) & 0.86 & 0.83 & 0.8 & 0.2 & 0.2 & 0.2 & 0.06 & 0.06 & 0.09 & 0.03 & 0.00 & 0.03 \\
 \hline
 Natural Language & 0.94 & 0.94 & 0.94 & 0.83 & 0.83 & 0.83 & .63 & 0.63 & 0.63 & 0.03 & 0.03 & 0.03 \\
 \hline
\end{tabular}
  \caption{Proportion of subjects including formulas $\phi_1 ... \phi_{12}$ from $\mathcal{D}_I$ across response types (see Technical Appendix C for the formulas). $\Phi_D$ was inferred from subject demonstrations $\mathcal{D}_D$. $\Phi_S$ is the human-provided PARCC specification. “Natural Language” refers to the human’s written specification. The proportion of subjects including each formula in their demonstration $\Phi_D$ is higher than either direct specification $\Phi_S$ or natural language. This supports hypothesis H4, and shows that inferring human specifications from demonstrations is more reliable than direct specification or language. }
\label{Specifications:table} 
\end{table*}

\begin{figure}
    \centering
    \includegraphics[width=\columnwidth]{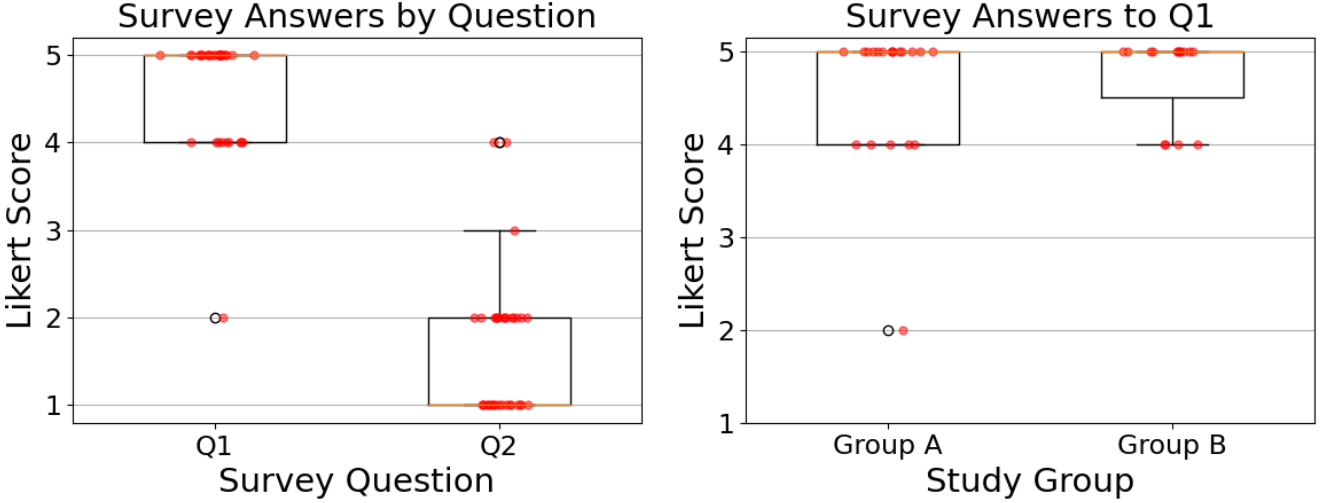}
    \caption{Box and whisker plots of Likert responses. (left) Likert responses to how well $C_D$ matched patterns in subjects' demonstrations (Q1), and how well $C_S$ matched patterns in subjects' demonstrations (Q2). Responses to Q1 were significantly greater than 3 ($p=5.1e-8$), and responses to Q2 were significantly less than 3 ($p=3.8e-7$). (right) Likert responses indicating how well $C_D$ matched patterns in subjects' demonstrations between Groups A and B. Responses did not differ significantly between the two groups.}%
    \label{fig:Boxplots}%
\end{figure}

Figure \ref{fig:Boxplots} and Table \ref{Specifications:table} summarize our major results. To characterize how successfully the inference procedure captured participants' intended specification across both groups, we determined via a Wilcoxon one-sample signed-rank test that responses to Q1 significantly exceeded 3 (i.e., that participants responded either ``agree" or ``strongly agree") ($p = 5.1\mathrm{e}{-8}$), implying that the inference algorithm captured the humans’ intended specifications, and supporting $H1$. 

Similarly, to characterize how successfully participants' provided specification $\Phi_S$ captured their own intended specification across both groups, a Wilcoxon one-sample signed-rank test also revealed that responses to Q2 were significantly less than 3 (i.e., that participants responded ``disagree" or ``strongly disagree") ($p = 3.8\mathrm{e}{-7}$), implying that humans' provided specifications differed from their intended specifications, and supporting $H2$. 

To determine whether the generation of non-specification data in Algorithm 2 is an appropriate analog for human demonstrations without prompting for a specification, we computed whether the responses to Q1 were significantly different between groups A and B via a Mann-Whitney U test. The results indicated no significant difference between the groups' responses ($p = 0.87$), supporting $H3$. 

To determine any differences between the human's provided specification ($\Phi_S$) and the inferred specification ($\Phi_D$), we compared these two specifications to the specification employed when creating the pre-generated ``initial" demonstrations $\mathcal{D}_I$ (Figure \ref{fig:packing_examples}). $\mathcal{D}_I$ used a specification of 12 PARCC formulas ($\phi_1 ... \phi_{12}$) in conjunction with each other.  These 12 formulas are described in Technical Appendix C. Table \ref{Specifications:table} shows the proportion of subjects for whom each of the 12 formulas appeared in $\Phi_D$, and the proportion of subjects who successfully encoded each formula in their specification $\Phi_S$. Across the 12 formulas, we used a two-sample Wilcoxon signed-rank test to characterize significant difference in the proportion of formulas specified by $\Phi_D$ and $\Phi_S$, and found that significantly fewer formulas were correctly specified by $\Phi_S$ compared with $\Phi_D$ ($p = 4.8\mathrm{e}{-4}$), supporting $H4$.

\subsection{Discussion}

The statistical support for $H1$ and $H2$ confirms that obtaining human PARCC specification via demonstrations provides an advantage over direct specification. Additionally, support for $H4$ suggests the perceived difference between $C_D$ and $C_S$ results from human misspecification, even when specification is performed via a formal language with inherent semantics translatable into natural language. This result is not surprising, given prior research demonstrates that humans often have difficulty interpreting and providing specifications in formal specification languages, with or without translation to natural language \cite{loomes1997formal,vinter1996seven,vinter1998evaluating,greenman2022little}. 

Our findings also suggest that human specification via natural language has shortfalls. Table \ref{Specifications:table} shows that subjects often underspecified when using natural language (though determining one-to-one correlations between natural language and PARCC specifications is subjective). Additionally, subjects' natural language often contain inconsistencies with regard to word choice; for example, some subjects described objects of a class ``clustering" around a corner of the box. However, some such participants always placed objects in contact with the walls comprising the corner, and some placed objects in the corner's vicinity. Such discrepancy shows natural language is an imprecise way to capture humans' intended specifications — and the underspecification of a human’s own intention indicates that there would be shortcomings to this approach even if the language were precise. 

Finally, support for $H3$ suggests that non-specification demonstrations for inference do not significantly vary between pseudo-randomly generated and human-generated data. Therefore, we conclude that pseudo-random object placement for non-specification demonstrations $\mathcal{R}$ (as in Algorithm \ref{InferAlg}) provides a reasonable analog for human-generated non-specification demonstrations. 


\section{Limitations and Future Work}
While PARCC and the inference method in this paper captures spatial specifications relevant to human demonstrators, there are several open areas for future work.  First, PARCC is limited to rectangular objects in two dimensions.  Extending to three dimensions and more varied geometries can improve applicability in situations where objects cannot be treated as bounding boxes in an image. However, considering a third dimension requires modeling object stacking and stability, which is outside this paper's scope. Second, the inference procedure relies on search over all possible disjunctive formulas allowed by a template.  While this works for a few object classes, it may be intractable for a larger number of classes due to combinatorial explosion.  In these cases, a sample based approach similar to Shah et al's use of Markov Chain Monte Carlo may be useful \cite{shah2018bayesian}.  Finally, inference over PARCC only considers rules that exist over all demonstrations.  In many situations, learning preferences in addition to strict rules may provide a route to capture more nuanced aspects of human demonstrations.  

\section{Conclusion}
In this work, we presented Positionally Augmented RCC, a specification language expressing spatial relationships between classes of objects. By utilizing RCC as the basis of our language, our method expresses human-intuitive spatial relationships between objects more easily than traditional spatial languages (e.g. STL). We also present an inference framework to learn PARCC specifications from demonstrations. Finally, via a human study, we show our framework's effectiveness in capturing human-intended spatial specifications and the advantage of learning-from-demonstration approaches to specification over direct human specification due to humans' tendency to mis- or under-specify. 

\section*{Acknowledgments}

DISTRIBUTION STATEMENT A. Approved for public re589 lease. Distribution is unlimited. This material is based upon work supported by the Under Secretary of Defense for Research and Engineering under Air Force Contract No. FA8702-15-D-0001. Any opinions, findings, conclusions or recommendations expressed in this material are those of the author(s) and do not necessarily reflect the views of the Under Secretary of Defense for Research and Engineering.

\bibliographystyle{named}
\bibliography{ijcai25}

@preamble{ "\newcommand{\noopsort}[1]{} "
	# "\newcommand{\printfirst}[2]{#1} "
	# "\newcommand{\singleletter}[1]{#1} "
	# "\newcommand{\switchargs}[2]{#2#1} " }

@ARTICLE{state-of-robots,
  title={The state of industrial robotics: Emerging technologies, challenges, and key research directions},
  author={Sanneman, Lindsay and Fourie, Christopher and Shah, Julie A},
  journal={arXiv preprint arXiv:2010.14537},
  year={2020}
}

@article{sun2022predicting,
  title={Predicting human discretion to adjust algorithmic prescription: A large-scale field experiment in warehouse operations},
  author={Sun, Jiankun and Zhang, Dennis J and Hu, Haoyuan and Van Mieghem, Jan A},
  journal={Management Science},
  volume={68},
  number={2},
  pages={846--865},
  year={2022},
  publisher={INFORMS}
}

@article{shah2018bayesian,
  title={Bayesian inference of temporal task specifications from demonstrations},
  author={Shah, Ankit and Kamath, Pritish and Shah, Julie A and Li, Shen},
  journal={Advances in Neural Information Processing Systems},
  volume={31},
  year={2018}
}

@inproceedings{haghighi2015spatel,
  title={SpaTeL: a novel spatial-temporal logic and its applications to networked systems},
  author={Haghighi, Iman and Jones, Austin and Kong, Zhaodan and Bartocci, Ezio and Gros, Radu and Belta, Calin},
  booktitle={Proceedings of the 18th International Conference on Hybrid Systems: Computation and Control},
  pages={189--198},
  year={2015}
}

@incollection{maler2004monitoring,
  title={Monitoring temporal properties of continuous signals},
  author={Maler, Oded and Nickovic, Dejan},
  booktitle={Formal Techniques, Modelling and Analysis of Timed and Fault-Tolerant Systems},
  pages={152--166},
  year={2004},
  publisher={Springer}
}

@article{randell1992spatial,
  title={A spatial logic based on regions and connection.},
  author={Randell, David A and Cui, Zhan and Cohn, Anthony G},
  journal={KR},
  volume={92},
  pages={165--176},
  year={1992}
}

@article{erbayrak2021multi,
  title={Multi-objective 3D bin packing problem with load balance and product family concerns},
  author={Erbayrak, Seda and {\"O}zk{\i}r, Vildan and Y{\i}ld{\i}r{\i}m, U Mahir},
  journal={Computers \& Industrial Engineering},
  volume={159},
  pages={107518},
  year={2021},
  publisher={Elsevier}
}

@inproceedings{edelkamp2014monte,
  title={Monte-Carlo tree search for 3D packing with object orientation},
  author={Edelkamp, Stefan and Gath, Max and Rohde, Moritz},
  booktitle={Joint German/Austrian Conference on Artificial Intelligence (K{\"u}nstliche Intelligenz)},
  pages={285--296},
  year={2014},
  organization={Springer}
}

@inproceedings{li2020stsl,
  title={STSL: A novel spatio-temporal specification language for cyber-physical systems},
  author={Li, Tengfei and Liu, Jing and Kang, JieXiang and Sun, Haiying and Yin, Wei and Chen, Xiaohong and Wang, Hui},
  booktitle={2020 IEEE 20th International Conference on Software Quality, Reliability and Security (QRS)},
  pages={309--319},
  year={2020},
  organization={IEEE}
}

@article{gross2016multi,
  title={Multi-modal referring expressions in human-human task descriptions and their implications for human-robot interaction},
  author={Gross, Stephanie and Krenn, Brigitte and Scheutz, Matthias},
  journal={Interaction Studies},
  volume={17},
  number={2},
  pages={180--210},
  year={2016},
  publisher={John Benjamins}
}

@inproceedings{nenzi2015qualitative,
  title={Qualitative and quantitative monitoring of spatio-temporal properties},
  author={Nenzi, Laura and Bortolussi, Luca and Ciancia, Vincenzo and Loreti, Michele and Massink, Mieke},
  booktitle={Runtime Verification: 6th International Conference, RV 2015, Vienna, Austria, September 22-25, 2015. Proceedings},
  pages={21--37},
  year={2015},
  organization={Springer}
}

@article{paul2018efficient,
  title={Efficient grounding of abstract spatial concepts for natural language interaction with robot platforms},
  author={Paul, Rohan and Arkin, Jacob and Aksaray, Derya and Roy, Nicholas and Howard, Thomas M},
  journal={The International Journal of Robotics Research},
  volume={37},
  number={10},
  pages={1269--1299},
  year={2018},
  publisher={SAGE Publications Sage UK: London, England}
}

@incollection{wiebrock2000inference,
  title={Inference and visualization of spatial relations},
  author={Wiebrock, Sylvia and Wittenburg, Lars and Schmid, Ute and Wysotzki, Fritz},
  booktitle={Spatial Cognition II},
  pages={212--224},
  year={2000},
  publisher={Springer}
}

@inproceedings{vasardani2013descriptions,
  title={From descriptions to depictions: A conceptual framework},
  author={Vasardani, Maria and Timpf, Sabine and Winter, Stephan and Tomko, Martin},
  booktitle={International Conference on Spatial Information Theory},
  pages={299--319},
  year={2013},
  organization={Springer}
}

@incollection{loomes1997formal,
  title={Formal methods: No cure for faulty reasoning},
  author={Loomes, Martin and Vinter, Rick},
  booktitle={Safer Systems},
  pages={67--78},
  year={1997},
  publisher={Springer}
}

@techreport{vinter1996seven,
  title={Seven lesser known myths of formal methods: uncovering the psychology of formal specification},
  author={Vinter, RJ and Loomes, MJ and Kornbrot, D},
  year={1996},
  institution={University of Hertfordshire}
}

@phdthesis{vinter1998evaluating,
  author      = {Vinter, RJ},
  title       = {Evaluating formal specifications: a cognitive approach},
  school      = {University of Hertfordshire},
  year        = {1998}
}

@article{greenman2022little,
  title={Little Tricky Logic: Misconceptions in the Understanding of {LTL}},
  author={Greenman, Ben and Saarinen, Sam and Nelson, Tim and Krishnamurthi, Shriram},
  journal={The Art, Science, and Engineering of Programming},
  volume={7},
  issue={2},
  year={2023}
}

@inproceedings{liu2022structformer,
  title={Structformer: Learning spatial structure for language-guided semantic rearrangement of novel objects},
  author={Liu, Weiyu and Paxton, Chris and Hermans, Tucker and Fox, Dieter},
  booktitle={2022 International Conference on Robotics and Automation (ICRA)},
  pages={6322--6329},
  year={2022},
  organization={IEEE}
}

@inproceedings{linard2020active,
  title={Active learning of signal temporal logic specifications},
  author={Linard, Alexis and Tumova, Jana},
  booktitle={2020 IEEE 16th International Conference on Automation Science and Engineering (CASE)},
  pages={779--785},
  year={2020},
  organization={IEEE}
}

@inproceedings{ma2020sastl,
  title={SaSTL: Spatial aggregation signal temporal logic for runtime monitoring in smart cities},
  author={Ma, Meiyi and Bartocci, Ezio and Lifland, Eli and Stankovic, John and Feng, Lu},
  booktitle={2020 ACM/IEEE 11th International Conference on Cyber-Physical Systems (ICCPS)},
  pages={51--62},
  year={2020},
  organization={IEEE}
}

@inproceedings{he2021know,
  title={Know your surroundings: Panoramic multi-object tracking by multimodality collaboration},
  author={He, Yuhang and Yu, Wentao and Han, Jie and Wei, Xing and Hong, Xiaopeng and Gong, Yihong},
  booktitle={Proceedings of the IEEE/CVF Conference on Computer Vision and Pattern Recognition},
  pages={2969--2980},
  year={2021}
}

@inproceedings{jia2021self,
  title={Self-Supervised Person Detection in 2D Range Data using a Calibrated Camera},
  author={Jia, Dan and Steinweg, Mats and Hermans, Alexander and Leibe, Bastian},
  booktitle={2021 IEEE International Conference on Robotics and Automation (ICRA)},
  pages={13301--13307},
  year={2021},
  organization={IEEE}
}

@inproceedings{van2016qualitative,
  title={Qualitative Privacy Description Language: Integrating Privacy Concepts, Languages, and Technologies},
  author={van de Ven, Jasper and Dylla, Frank},
  booktitle={Privacy Technologies and Policy: 4th Annual Privacy Forum, APF 2016, Frankfurt/Main, Germany, September 7-8, 2016, Proceedings 4},
  pages={171--189},
  year={2016},
  organization={Springer}
}

@inproceedings{dwyer1999patterns,
  title={Patterns in property specifications for finite-state verification},
  author={Dwyer, Matthew B and Avrunin, George S and Corbett, James C},
  booktitle={Proceedings of the 21st international conference on Software engineering},
  pages={411--420},
  year={1999}
}

@article{vazquez2018learning,
  title={Learning task specifications from demonstrations},
  author={Vazquez-Chanlatte, Marcell and Jha, Susmit and Tiwari, Ashish and Ho, Mark K and Seshia, Sanjit},
  journal={Advances in neural information processing systems},
  volume={31},
  year={2018}
}

@article{xu2016temporal,
  title={A temporal logic inference approach for model discrimination},
  author={Xu, Zhe and Birtwistle, Marc and Belta, Calin and Julius, Agung},
  journal={IEEE life sciences letters},
  volume={2},
  number={3},
  pages={19--22},
  year={2016},
  publisher={IEEE}
}

@inproceedings{jin2013mining,
  title={Mining requirements from closed-loop control models},
  author={Jin, Xiaoqing and Donz{\'e}, Alexandre and Deshmukh, Jyotirmoy V and Seshia, Sanjit A},
  booktitle={Proceedings of the 16th international conference on Hybrid systems: computation and control},
  pages={43--52},
  year={2013}
}

@article{hoxha2018mining,
  title={Mining parametric temporal logic properties in model-based design for cyber-physical systems},
  author={Hoxha, Bardh and Dokhanchi, Adel and Fainekos, Georgios},
  journal={International Journal on Software Tools for Technology Transfer},
  volume={20},
  pages={79--93},
  year={2018},
  publisher={Springer}
}

@article{zendehdel2023real,
  title={Real-time tool detection in smart manufacturing using You-Only-Look-Once (YOLO) v5},
  author={Zendehdel, Niloofar and Chen, Haodong and Leu, Ming C},
  journal={Manufacturing Letters},
  volume={35},
  pages={1052--1059},
  year={2023},
  publisher={Elsevier}
}

@article{ali2024yolo,
  title={The YOLO framework: A comprehensive review of evolution, applications, and benchmarks in object detection},
  author={Ali, Momina Liaqat and Zhang, Zhou},
  journal={Computers},
  volume={13},
  number={12},
  pages={336},
  year={2024},
  publisher={MDPI}
}



\end{document}


\maketitle

\appendix
\section{Relations in RCC}
Table \ref{RCC:table} describes the 10 canonical RCC relations \cite{randell1992spatial}.  The fragment used in PARCC ($EC$, externally connected to, and $DR$, discrete from) is bolded. We do not include $P$, $PP$, $EQ$, $O$, $PO$, $TPP$, and $NTPP$ in PARCC because they describe some form of overlap between regions, which we specifically disallow given that our regions represent physical objects. We also do not include $DC$ (disconnected from), because it can already be easily described by our fragment via $DR(x, y) \land \lnot EC(x, y)$.

\begin{table*}
\centering
 \caption{Description of relations included in the region connection calculus. (The fragment used in PARCC is bolded.)}
\begin{tabular}{ |p{2.1cm}|p{5.5cm}|p{6.5cm}| }
 \hline
 \multicolumn{3}{|c|}{RCC Relations} \\
 \hline
 Relation & Interpretation & Definition\\
 \hline
 $DC(x,y)$   & $x$ is disconnected from $y$                 & $\lnot C(x, y)$\\
 $P(x,y)$    & $x$ is a part of $y$                         & $\forall z [C(z,x) \rightarrow C(z,y)]$\\
 $PP(x,y)$   & $x$ is a proper part of $y$                  & $P(x,y) \land \lnot P(y,x)$\\
 $EQ(x,y)$   & $x$ is identical with $y$                    & $P(x,y) \land P(y,x)$\\
 $O(x,y)$    & $x$ overlaps $y$                             & $\exists z [P(z,x) \land P(z,y)]$\\
 $\mathbf{DR(x,y)}$   & $x$ is discrete from $y$                     & $\lnot O(x,y)$ \\
 $PO(x,y)$   & $x$ partially overlaps $y$                   & $O(x, y) \land \lnot P(x, y) \land \lnot P(y, x)$ \\
 $\mathbf{EC(x,y)}$   & $x$ is externally connected to $y$           & $C(x, y) \land \lnot O(x,y)$\\
 $TPP(x,y)$  & $x$ is a tangential proper part of $y$       & $PP(x,y) \land \exists z [EC(z,x) \land EC(z,y)]$\\
 $NTPP(x,y)$ & $x$ is a non-tangential proper part of $y$   & $PP(x,y) \land \lnot \exists z [EC(z,x) \land EC(z,y)]$\\
 \hline
\end{tabular}
\label{RCC:table} 
\end{table*}

\section{Comparison to STL Spatial Specification}

Prior research has introduced languages to encode spatial specifications in formal logic via Signal Temporal Logic \cite{maler2004monitoring,linard2020active} and quad-tree representations \cite{haghighi2015spatel}.  In this appendix, we show the benefits of using PARCC over these methods for ease of use and interpretability for describing object relations. At a high level, the languages differ because PARCC operates at a qualitative level that exposes semantics that are meaningful to humans, using concepts related to whether or not objects are touching, and directionality.  In contrast, STL is a quantitative specification, and SpaTeL is a quad-tree representation, neither of which is easy for a human to parse.

STL and its variants are highly expressive and capable of capturing the relations described in PARCC.  However, given that they are not designed to encode spatial relationships between objects, the notation is generally clumsy.  To depict the difference, we express the two primary PARCC relationships ($EC_i$ and $DR_i$) to in STL. In all the following STL examples, the signal includes the $x$ and $y$ position of all objects involved. 

First, we consider the simple PARCC specification $EC_N(A, B)$.  This specification indicates that for every object $a$ in class $A$, there is some object $b$ in class $B$ such that they overlap in the east/west direction and precisely meet in the north/south direction: 

\begin{multline}
    EC_N(A,B) = G_{[0,\infty]} ( \underset{a \in A}{\bigwedge} \underset{b \in B}{\bigvee} ( \lnot (b_y - \frac{b_w}{2} > a_y + \frac{a_w}{2}) \\
    \wedge \lnot (b_y - \frac{b_w}{2} < a_y + \frac{a_w}{2}) \\ 
    \wedge (b_x - \frac{b_l}{2} < a_x + \frac{a_l}{2}) \\
    \wedge (b_x + \frac{b_l}{2} < a_x - \frac{a_l}{2}) ))
\end{multline}

Where $G_{[0,\infty]}$ notates the temporal logic "Globally", indicating that the specification must remain true for all time.  Second, we express the PARCC specification $DR_N(A, B)$:

\begin{multline}
    DR_N(A,B) = G_{[0,\infty]} ( \underset{a \in A}{\bigwedge} \underset{b \in B}{\bigwedge} (a_y - \frac{a_w}{2} > b_y + \frac{b_w}{2}))
\end{multline}

Notice that in both circumstances, STL would be far more clumsy for operators and much more complex to extract from demonstrations if attempting to infer a specification. Additionally, STL's temporal requirements (i.e. the start and end times, 0 and $\infty$) are extraneous for static specifications like the object placement task.

Specifications via quad-tree representations such as SpaTeL are even less equipped to handle the qualitative relations provided by RCC and PARCC \cite{haghighi2015spatel}.  Consider trying to encode an $EC_N$ relation via quad-trees.  Assuming we model an object's placement as the existence of the object's position $(o_x, o_y)$ in a node of the tree, ensuring that two objects touch exactly would require a theoretically infinite tree to position the two exactly and encoding the distance relationships between two objects within such a logical framework would be difficult.  Even if we discretized the space, the number of spatial partitions to approximate contact between every two objects in a class would be high and nearly uninterpretable.  Furthermore, attempting to infer such a specification or satisfy a specification with object placements would likely be intractable.  

\section{Intended Specification for User Study}

Below are the 12 PARCC formulae which together make the specification used to create the ``initial" demonstrations $\mathcal{D}_I$ in our user study.  

\begin{enumerate}
    \item $DR_N(B, R)$ -- Blue objects are North of Red objects.  
    \item $DR_E(G, R)$ -- Green objects are East of Red objects.
    \item $DR_E(G, B)$ -- Green objects are East of Blue objects.  
    \item $\underset{i \in \{N, S, E, W\}}{\bigvee} EC_i(B, B)$ -- Blue objects are externally connected to another blue object on any side.  
    \item $\underset{i \in \{N, S, E, W\}}{\bigvee} EC_i(R, R)$ -- Red objects are externally connected to another red object on any side.  
    \item $\underset{i \in \{N, S, E, W\}}{\bigvee} EC_i(G, G)$ -- Green objects are externally connected to another green object on any side.  
    \item $(EC_W(B, B) \vee EC_W(B, WA)) \bigwedge (\underset{i \in \{N, S\}}{\bigvee} EC_i(B, B) \vee EC_i(B, WA) \vee EC_i(B, R))$ -- Blue objects must be in contact with a Blue or Wall object on the west side and a Blue, Red, or Wall object on the north or south sides. This results in the object existing in the ``corner" of at least 2 other objects.  
    \item $(EC_W(R, R) \vee EC_W(R, WA)) \bigwedge (\underset{i \in \{N, S\}}{\bigvee} EC_i(R, B) \vee EC_i(R, WA) \vee EC_i(R, R))$ -- Red objects must be in contact with a Red or Wall object on the west side and a Blue, Red, or Wall object on the north or south sides. This results in the object existing in the ``corner" of at least 2 other objects.  
    \item $(EC_E(G, G) \vee EC_E(G, WA)) \bigwedge (\underset{i \in \{N, S\}}{\bigvee} EC_i(G, G) \vee EC_i(G, WA))$ -- Green objects must be in contact with a Green or Wall object on the east side and a Green or Wall object on the north or south sides. This results in the object existing in the ``corner" of at least 2 other objects.  
    \item $\underset{i \in \{N, S, E, W\}}{\bigwedge} \; \lnot DR_i(B, W)  $ -- Blue objects are not North, South, East, or West of all wall objects (i.e. inside the box). 
    \item $\underset{i \in \{N, S, E, W\}}{\bigwedge} \; \lnot DR_i(R, W)  $ -- Red objects are not North, South, East, or West of all wall objects (i.e. inside the box). 
    \item $\underset{i \in \{N, S, E, W\}}{\bigwedge} \; \lnot DR_i(G, W)  $ -- Green objects are not North, South, East, or West of all wall objects (i.e. inside the box). 
\end{enumerate}

\section{Comparison of PARCC Templates}

The inference procedure proposed for PARCC in the main paper allows use of a ``template" to limit the search space to the most relevant disjunctive formulas.  Similar templates have been used in other specification languages.  Dwyer et al, for example identified templates for LTL most relevant to modeling real-world processes \cite{dwyer1999patterns}. Others have used templating in domains ranging from control systems to inferring levels of protein expression \cite{jin2013mining,hoxha2018mining,xu2016temporal}.  The template we choose for our study is intended to similarly represent the most relevant part of the search space for human demonstrators in general.  While a full examination of possible templates deserves its own work, in this appendix we compare the inferred specification from human subject demonstrations for three choices of template.  

First, we use the template from the human study described in section 5.2 (which we will refer to as ``original").  Second, we use a template that requires disjunctive formulas to contain either EC or DC class relations, but not both.  Notice that the formulas allowed by this template are a superset of the original template, and therefore we refer to it as the ``relaxed" template.  Third, we use a template that imposes all restrictions from the original template in the human study, and in addition disallows all $DR_N$ and $DR_S$ relations.  The formulas allowed by this template are a subset of the original template, and therefore we refer to it as the ``restrictive" template.  We include this template as an example of the restrictions an operator can make to ensure that rules which are not desired in a specification (but may arise in demonstrations) can be eliminated.  

\begin{table}
\centering
 \caption{Intersection over union comparison of disjunctions inferred from human demonstrations in the main paper using the original template versus more relaxed and restrictive templates. Especially take note that the intersection over union of 1 for the relaxed template indicates that (despite allowing a larger set of possible disjunctions) the relaxed template inferred the exact specification as the original template. This shows that our choice of template for the study imposed minimal inductive bias on the inferred specifications.}
\begin{tabular}{ |C{2.1cm}|C{2 cm}|C{2 cm}| }
 \hline
 \multicolumn{3}{|c|}{Inference Comparison to Original Template} \\
 \Xhline{3\arrayrulewidth}
 \multirow{2}{*}{Template} & \multicolumn{2}{|c|}{Intersection Over Union} \\
 \cline{2-3}
 & Mean & SD \\
 \Xhline{3\arrayrulewidth}
 Relaxed   & 1     & 0    \\ \hline
 Restrictive & .76 & .074 \\
 \hline
\end{tabular}
\label{table:IntersectionOverUnion} 
\end{table}

\begin{table}
\centering
 \caption{Average runtime to infer specifications from each participants' demonstrations in the human study for each template. We include both the runtime in seconds and the number of formulas formulas that the algorithm checked against demonstrations for satisfaction. }
\begin{tabular}{ |C{1.5cm}|C{1 cm}|C{1cm}|C{1cm}|C{1cm}| }
 \hline
 \multicolumn{5}{|c|}{Runtime Complexity} \\
 \Xhline{3\arrayrulewidth}
 \multirow{2}{*}{Template} &  \multicolumn{2}{|c|}{Runtime (s)} & \multicolumn{2}{|c|}{\# Formulas} \\ 
  \cline{2-5}
  & Mean        & SD & Mean        & SD \\
 \Xhline{3\arrayrulewidth}
 Original    & 4.46  & .47  & 48967  &  2640  \\ \hline
 Relaxed     & 60.63 & 6.61 & 470465 &  29876 \\ \hline
 Restrictive & 2.56  & 0.39 & 29559  &  2012  \\ \hline
\end{tabular}
\label{table:Runtime} 
\end{table}

Table \ref{table:IntersectionOverUnion} compares the set of disjunctive formulas inferred using the restrictive and relaxed templates to those inferred using the original template.  For a similarity measure, we use the average intersection over union of formulas inferred from each participants' demonstrations in the human study ($\mathcal{D}_D$) across the three templates.  Notice that the intersection over union between the relaxed and original templates are 1.  Therefore, despite the relaxed template allowing a larger set of formulas, the inference procedure using the original and relaxed templates always inferred the same exact specification.  This result provides evidence that our template imposed minimal inductive bias on the specifications inferred during the human study, and that the original template captures the types of spatial constraints likely to be considered and reproduced by people. The intersection over union between the original and restrictive templates are .76, showing that the template successfully removed a set of formulas that an operator may not have wanted. 

Table \ref{table:Runtime} compares the complexity of inference for each template, both in terms of time and total number of disjunctive formulas the algorithm checked against demonstrations for satisfaction.  These inference times are once again averaged across the sets of demonstrations $\mathcal{D}_D$ in the human study.  Note that the number of formulas checked may change between different participants' demonstrations since we avoid checking a disjunction that is trivially satisfied by another disjunction already known to satisfy demonstrations (e.g. if we know $EC_N(R, G)$ satisfies all demonstrations, we do not check $EC_N(R, G) \vee EC_N(R, B)$). Notice that the relaxed template checked nearly 10 times the formulas and required 15 times the runtime as the original template, despite the two templates ultimately inferring the exact same specification for every set of demonstrations. This boost in time without change in inferred formulas demonstrates the need for a template: if a template can capture information common to human-intuitive spatial relationships, utilizing this structure can allow significant increases in efficiency without sacrificing performance. Additionally, the lack of difference in inferred formulas between the original and relaxed templates provides evidence that the original template indeed captures aspects of human-intuitive object placement constraints, though a full validation of this claim is outside the scope of this paper. Finally, the restricted template required around half the number of formulas and runtime compared to the original.  This shows that if a set of formulas are known to be irrelevant, using a template to remove them from the search space can be beneficial for runtime.

\bibliographystyle{named}
\bibliography{ijcai25}